\documentclass[conference]{IEEEtran}
\IEEEoverridecommandlockouts
\usepackage{cite}
\usepackage{amsmath,amssymb,amsfonts}
\usepackage{algorithmic}
\usepackage{graphicx}
\usepackage{textcomp}
\usepackage{xcolor}
\usepackage{bm}
\usepackage{subfigure}
\usepackage{booktabs}
\usepackage{caption}
\usepackage{multirow}
\def\BibTeX{{\rm B\kern-.05em{\sc i\kern-.025em b}\kern-.08em
    T\kern-.1667em\lower.7ex\hbox{E}\kern-.125emX}}

\newcommand{\cE}{\mathcal{E}}

\newcommand{\cG}{\mathcal{G}}

\newcommand{\cL}{\mathcal{L}}

\newcommand{\cN}{\mathcal{N}}
\newcommand{\cO}{\mathcal{O}}

\newcommand{\cR}{\mathcal{R}}

\newcommand{\cV}{\mathcal{V}}

\usepackage{graphicx}
\usepackage{soul}
\begin{document}

\title{Heterogeneous Molecular Graph Neural Networks for Predicting Molecule Properties
}

% \author{\IEEEauthorblockN{Anonymous}}
\author{
\IEEEauthorblockN{Zeren Shui}
\IEEEauthorblockA{Computer Science \& Engineering\\
University of Minnesota\\
Minneapolis, USA \\
shuix007@umn.edu}
\and
\IEEEauthorblockN{George Karypis}
\IEEEauthorblockA{Computer Science \& Engineering \\
University of Minnesota\\
Minneapolis, USA \\
karypis@umn.edu}
}

\maketitle

\begin{abstract}
As they carry great potential for modeling complex interactions, graph neural network (GNN)-based methods have been widely used to predict quantum mechanical properties of molecules. Most of the existing methods treat molecules as molecular graphs in which atoms are modeled as nodes. They characterize each atom's chemical environment by modeling its pairwise interactions with other atoms in the molecule. Although these methods achieve a great success, limited amount of works explicitly take many-body interactions, i.e., interactions between three and more atoms, into consideration. In this paper, we introduce a novel graph representation of molecules, heterogeneous molecular graph (HMG) in which nodes and edges are of various types, to model many-body interactions. HMGs have the potential to carry complex geometric information. To leverage the rich information stored in HMGs for chemical prediction problems, we build heterogeneous molecular graph neural networks (HMGNN) on the basis of a neural message passing scheme. HMGNN incorporates global molecule representations and an attention mechanism into the prediction process. The predictions of HMGNN are invariant to translation and rotation of atom coordinates, and permutation of atom indices. Our model achieves state-of-the-art performance in 9 out of 12 tasks on the QM9 dataset.
\end{abstract}


\section{Introduction}
Predicting quantum mechanical properties of molecules based on their structures is important for molecule screening and drug design. We can compute exact molecular properties by solving the many-body Schr\"{o}dinger equation. However, closed form solution to this equation is only available for simple systems. Although researchers developed methods such as Density Functional Theory (DFT) \cite{hohenberg1964inhomogeneous} to approximate the solution, the computational cost of these methods scales poorly and is worse than $\cO(n^3)$ w.r.t. the number of electrons. 

Recently, researchers have been developing machine learning methods that are orders of magnitude faster with a moderate compromise in prediction accuracy. Among the machine learning approaches, graph neural network (GNN)-based methods attract a lot of research attention as their ability to model complex interactions among atoms. These methods treat molecules as molecular graphs (e.g., distance graphs \cite{unke2019physnet, SchNet-1, SchNet-2, MGCN}, chemical graphs \cite{MPNN}, $K$-nearest neighbor graphs \cite{NMP-edge}) in which atoms are modeled as nodes. They compute an atom's low-dimensional representation as a function of its feature and characteristics of its graph neighbors. The low-dimensional representations are then used to estimate the local contribution of the atoms to the desired property, or to compute a global representation of the molecule for downstream predictions.

% The hierarchical structure of GNNs is analogous to an important scheme for modeling molecular energy surfaces, the many-body expansion (MBE). The MBE assumes the energy of a system can be decomposed as the summation of contributions of many-bodies. An $n$-body (the value of $n$ is called the \emph{order} of the many-body expansion) is a group of $n$ particles that functions as a whole entity. To model the MBE approach, researchers have built GNN-based methods that combine the contributions of different order MBEs when computing atom-wise predictions.
%Researchers have built GNN-based methods that decompose atom-wise predictions into contributions over many-body expansions of different order by computing outputs after each GNN layer. 

% These methods compute each atom's contribution to the desired property from its atom embedding and aggregate the atom-wise contributions as the final prediction.

The many-body expansion (MBE)~\cite{MBE-1, MBE-2, MBE-NN} is an important scheme that computes the energy of an $N$-particle system as the sum of the contributions of many-body terms
\begin{equation} \label{eq:MBE}
    E = \sum_{i} E_i + \sum_{i < j} E_{ij} + \sum_{i < j < k} E_{ijk} + \cdots + E_{12\cdots N},
\end{equation}
where $E_i$ is the local energy contribution of a single atom, $E_{ij}$ is the energy contribution of a two-body (a group of two atoms), $E_{ijk}$ is the energy contribution of a three-body, and eventually $E_{12\cdots N}$ is the contribution of the body formed by all the atoms in the molecule. Note that, the local contribution to the total energy decreases fast with the number of atoms in the many-body. 
As most of the existing GNN-based methods are developed on molecular graphs, they focus mainly on modeling atom-based representations, interactions, and predictions which correspond to the first two terms of the series and do not have an explicit characterization of the higher order terms. This may compromise their accuracy in the chemical prediction problems.

In this paper, we introduce a novel graph representation of molecules, \emph{heterogeneous molecular graph} (HMG), to explicitly model many-body interactions. A $p$-body (the value of $p$ is called the \emph{order} of the many-body) is a group of $p$ atoms that functions as a whole entity. In HMGs, a $p$-body is modeled as a node of order $p$. Nodes connect to nodes of the same or different order via different types of edges. This heterogeneous structure allows us to explicitly model interactions, representations, and predictions associated with many-bodies. Moreover, edges between nodes of the same order carry the potential of incorporating complex geometric information (e.g., bond angles and dihedral angles) into node embeddings. 

To leverage the rich information stored in HMG for tasks of molecular property predictions, we design heterogeneous molecular graph neural networks (HMGNN) by following a message passing framework. In the message passing framework \cite{MPNN}, nodes send and receive messages from their neighbors and update their low-dimensional representations using the received messages. HMGNN is a multi-task learning \cite{MTL} model whose design is inspired by the MBE of energy surfaces. In HMGNN, each many-body order possesses its own set of parameters and shares computations with other orders. In the prediction phase, HMGNN computes one estimation for each many-body and aggregates them based on their orders. It uses an attention-based model that takes into account a global representation of the molecule to fuse the prediction of different orders, which correspond to different terms in Eq~\ref{eq:MBE}. We design a multi-task learning loss that enforces the prediction of each order and the fused prediction to be close to the true target. Experimental results show that the fused prediction is better than any of the standalone predictions. The fusing weight of the predictions are also consistent with the convergence assumption in the many-body expansion. 

The main contribution of this work lies in two folds. First, we propose HMG which allows graph learning methods to explicitly model many-body representation, interaction, and prediction. Second, we develop a multi-task learning method HMGNN for the task of molecule property prediction. HMGNN explicitly incorporates many-body interaction and a global molecule representation into the prediction process and achieves state-of-the-art performance on the QM9 dataset \cite{QM9-1, QM9-2}. The code of HMGNN is available online\footnote{https://github.com/shuix007/HMGNN}.

\section{Review of relevant prior works}

Traditionally, prediction of many important molecular properties such as atomization energies relies on methods that approximate the solution of the many-body Schr\"{o}dinger equation such as density function theory (DFT) and its variants \cite{DFT}. This class of methods involves solving complex linear systems and has a computational complexity worse than $\cO(n^3)$ where $n$ is the number of atoms. 

Recent years have seen a surge in data-driven methods that train machine learning models to learn patterns from molecule databases. The learned patterns are assumed to be general in chemical space and can be used to estimate properties of unknown compounds. These attempts started from \cite{faber2017machine, bartok2017machine} which feed hand-crafted molecule descriptors (e.g., Coulomb matrix, bag of bonds) into regression models such as linear regression and random forests. These methods rely heavily on the quality of the crafted descriptors and have limited representation power. 

Recently, graph neural networks (GNN) have been achieving a great success in graph-related applications \cite{GCN, GAT, SAGE, DiffPool}. In chemistry, researchers developed GNN-based method for learning tasks over graph represented molecules. The authors of \cite{MPNN} introduced a generic framework over chemical graphs that models interactions between atoms in a message passing fashion. In \cite{DTNN, SchNet-1, SchNet-2}, the authors designed neural network structures that have no dependency on hand-crafted features but learn molecule representations from only atom types and coordinates. Since GNNs possess a hierarchical structure, i.e., they iteratively apply GNN layers on graphs to encode each node's multi-hop neighbors into its embedding, GNN-based methods \cite{HIP-NN} and \cite{unke2019physnet} further decompose atom-wise prediction to layer-wise atom prediction to fit in the MBE framework. Although these methods include many-body contributions into final predictions, they do not have an explicit modeling of many-body representations and interactions. Some recent works have incorporated many-body interactions and representations by updating edge embeddings along message passing \cite{NMP-edge} or by passing messages on line graphs of the corresponding molecular graphs \cite{DimeNet}. However, these methods capture only partial many-body interactions and lack many-body predictions.

Equivariant neural network is another class of neural network methods that has been applied in chemical prediction problems. The notion of group equivariant neural network was first introduced by \cite{cohen2016group} in the domain of image processing. Later, researchers developed neural network methods that are equivariant to continuous rotations for learning representations for 3D objects, including molecules \cite{anderson2019cormorant, thomas2018tensor, kondor2018covariant}. These methods achieve rotation invariance by transforming objects from Euclidean space to Fourier space and conducting computations in Fourier space. In these methods, each many-body interacts only with itself but not other many-bodies. Thus, they are not optimal in predicting molecule properties.

% \begin{figure*}[t]
% % \begin{tabular}{cc}
% \centering
% \subfigure[A methanol molecule.]{
% \includegraphics[width=0.3\textwidth]{ch3oh.pdf}
% \label{subfig:molecule}}
% \subfigure[A molecular graph of the methanol molecule.]{
% \includegraphics[width=0.3\textwidth]{ch3oh_MG.pdf}
% \label{subfig:mg}}
% \\\vspace{.2cm}
% \subfigure[A heterogeneous molecular graph of the molecular graph.]{
% \includegraphics[width=0.4\textwidth]{ch3oh_HMG.pdf}
% \label{subfig:hmg}}

\begin{figure*}[t]
% \begin{tabular}{cc}
\centering
\subfigure[A formaldehyde molecule.]{
\includegraphics[width=0.2\textwidth]{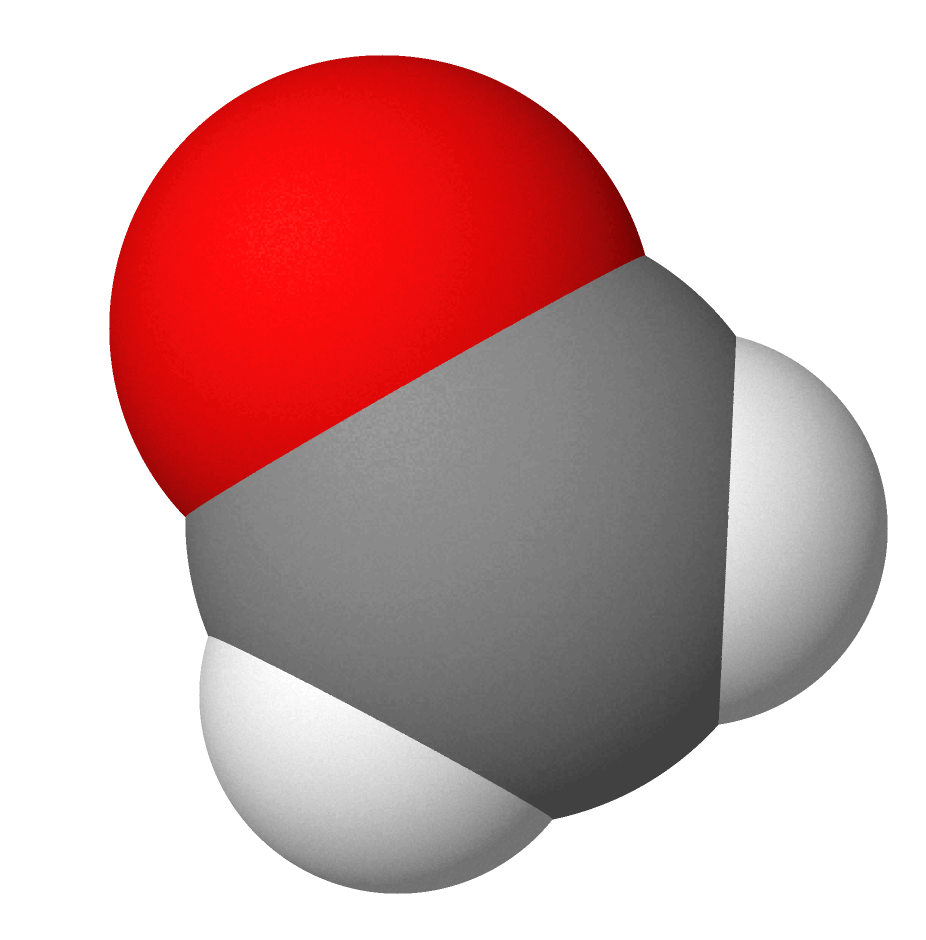}
\label{subfig:molecule}}
\subfigure[A molecular graph of the formaldehyde molecule.]{
\includegraphics[width=0.3\textwidth]{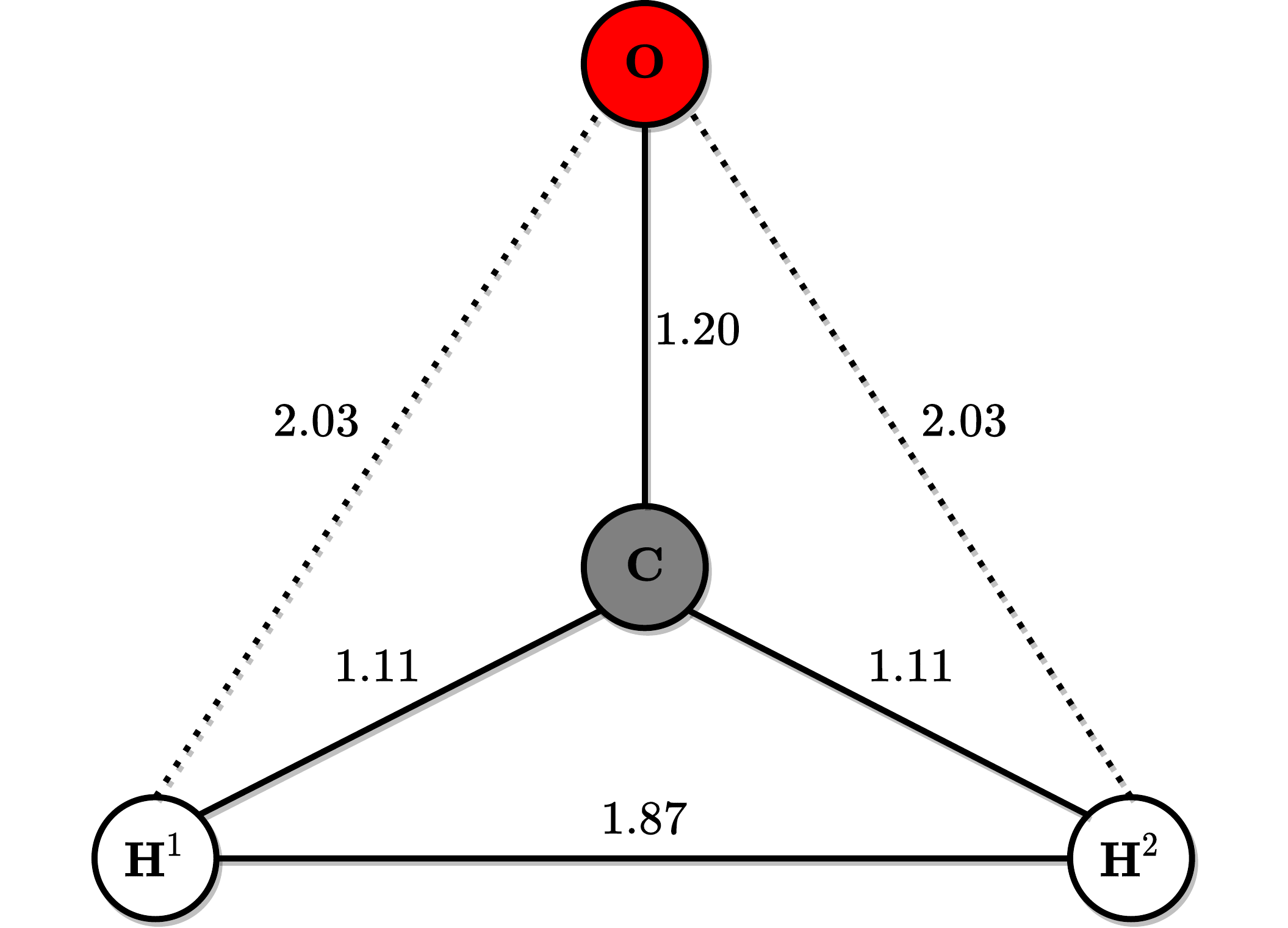}
\label{subfig:mg}}
% \\\vspace{.2cm}
\subfigure[A heterogeneous molecular graph of the molecular graph.]{
\includegraphics[width=0.35\textwidth]{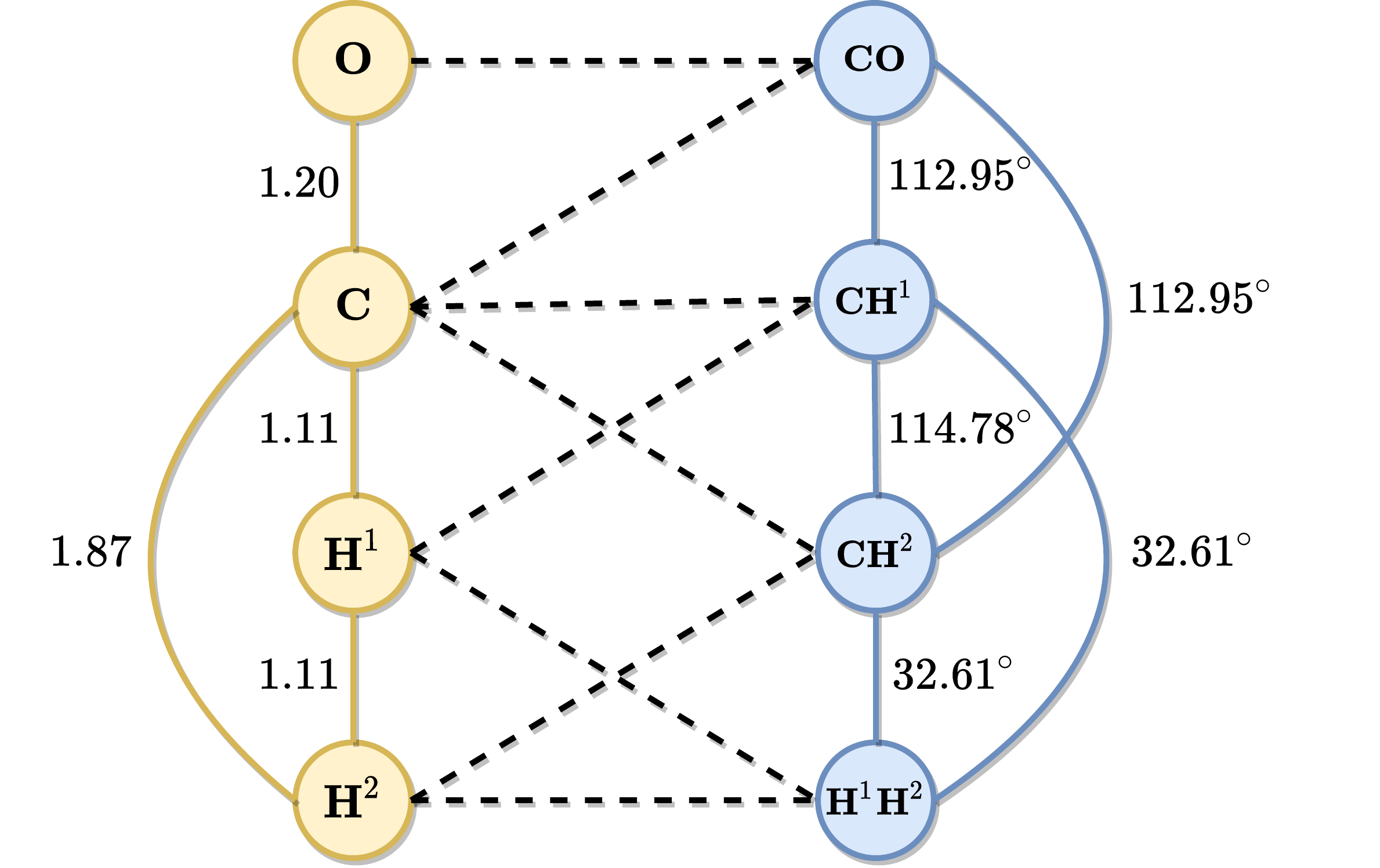}
\label{subfig:hmg}}

\caption{An example of heterogeneous molecular graph (HMG). Figure \ref{subfig:molecule} is a spatial structure of a formaldehyde ($\text{CH}_2\text{O}$) molecule. Each atom in the molecule is associated with a three-dimensional coordinates in the Euclidean space. Figure \ref{subfig:mg} is the molecular graph of the methanol molecule with a cutoff distance $c=2$. We convert atom coordinates to pair-wise distances to guarantee translation and rotation invariance of the representation. We denote edges whose distances are less than $c$ using black solid lines, and edges that are broke by the cutoff using black dotted lines. Figure \ref{subfig:hmg} is a HMG of order two constructed from the molecular graph. There are two types of nodes ($1$-bodies and $2$-bodies denoted by yellow and blue circles, respectively) and three types of edges ($1$-$1$ and $2$-$2$ denoted by yellow and blue lines, respectively, $1$-$2$ denoted by black dashed lines) in the HMG. Edges between nodes of the same order are associated with features that depict the geometric relation between the nodes (distance for $1$-$1$ edges, angle for $2$-$2$ edges). }
\label{fig:HMG}
\end{figure*}

\section{Notations and Definitions}
We denote matrices by bold upper-case letters (e.g., $\mathbf{W}$), and vectors by bold lower-case letters (e.g., $\mathbf{x}$). We denote entries of a matrix/vector by lower-case letter with subscripts (e.g., $x_{ij}$/$x_{i}$). We use superscripts to indicate variables at the $t$-th message passing layer (e.g., $\mathbf{h}^{(t)}$). We denote \emph{molecular graphs} by $\cG = \left(\cV, \cE\right)$ where $\cV$ and $\cE$ represent the set of nodes (atoms) and edges, respectively. Two atoms are connected in a molecular graph when the Euclidean distance between them is less than a cutoff threshold $c > 0$. Each edge in the graph is associated with a distance to store the geometric structure of the molecule. We define a $p$-body in a molecular graph $\cG$ as a $p$-clique of the graph. We refer to the value of $p$ as the order of the many-body.

% In a molecule, we define a $p$-body as a $p$-clique of 

% In a molecule, we define a $p$-body as a complete graph constructed from a subset of $p$ atoms of the molecule. We refer to the value of $p$ as the order of the many-body. Each edge in the graph is associated with a distance to capture the geometric structure of the body.

\section{Heterogeneous Molecular Graph and Many-Body Interactions}

In this section, we illustrate the construction of \emph{heterogeneous molecular graphs} (HMG) and how we leverage the heterogeneous structure of HMGs to model many-body representations and interactions. 

\subsection{Heterogeneous Molecular Graph} \label{Sec:HMG}

An HMG is a graph in which nodes are many-bodies and edges are defined by various types of geometric and set relations. HMGs are constructed from molecular graphs. We denote an HMG of order $N$ of a molecular graph $\cG$ as $H_N(\cG) = \left(\left\{\cV_p\right\}, \left\{\cE_{pq}\right\} \right)$ where $1 \leq p \leq q \leq N$, $\cV_p$ is the set of $p$-bodies in $\cG$ (i.e., all $p$-cliques of $\cG$), and $\cE_{pq}$ is the set of edges between $\cV_p$ and $\cV_q$. We denote the order $p$ of $p$-bodies as the node type and $p$-$q$ as the type of the edges that connect nodes of order $p$ and nodes of order $q$. Given two nodes $i \in \cV_p$ and $j \in \cV_q$, when they are of the same order, i.e., $p=q$, $i$ and $j$ are connected if they share $p-1$ atoms. A special case is when $p=q=1$, instead of building a complete graph, we use the edge set $\cE$ of the molecular graph to define connections. When the two nodes are of different orders, presumably $p < q$, $(i, j) \in \cE_{pq}$ if $i$ is a sub-graph of $j$. An example HMG is shown in Figure \ref{fig:HMG}. With this formulation, we can explicitly model up to $N$-body representations by node embeddings and $N+1$-body interactions by message passing.  

In an HMG, each node $i$ of order $p$ is associated with a discrete feature $Z_{p, i}$ that indicates its atomic composition, and a continuous feature $\mathbf{x}_{p, i}$ that describes aspects of its geometry. Note that, nodes of order $1$ do not have continuous features since they are points in the Euclidean space and do not have geometric structure. Each edge $(i, j)$ is associated with an edge feature $\mathbf{e}_{p, ij}$ when $i$ and $j$ are of the same order $p$. The edge feature characterizes the geometric relation between the two nodes, e.g., distance between atoms, angles between bonds. In this paper, we use a hash function to map the set of atomic numbers of the atoms to $Z_{p, i}$. Construction of continuous node features and edge features requires feature engineering especially when order of the many-bodies are high. We will illustrate how we convert geometric information to feature vectors up to the second order in Section \ref{Sec:implement}.

\subsection{Message Passing on Heterogeneous Molecular Graphs} 
The message passing framework consists of two phases, message passing and node update. On molecular graphs, each node (atom) $i$ sends/receives messages to/from its neighbors and uses the received messages to update its embedding
\begin{equation} \label{eq:MP}
\begin{gathered}
    \mathbf{m}_i^{(t)} = \sum_{j \in \cN(i)} f^{(t)} \left(\mathbf{h}_i^{(t)}, \mathbf{h}_j^{(t)}, \mathbf{e}_{ij}\right) \\
    \mathbf{h}_i^{(t + 1)} = g^{(t)} \left(\mathbf{h}_i^{(t)}, \mathbf{m}_i^{(t)}\right).
\end{gathered}
\end{equation}
In Eq-\ref{eq:MP}, $\cN(i)$ is the set of neighbor nodes of $i$, $\mathbf{h}_i^{(t)}$ is the node (atom) embedding of $i$, $\mathbf{m}_i^{(t)}$ is the aggregation of messages from $i$'s neighbor nodes,  $\mathbf{e}_{ij}$ is the edge feature associated with the edge between $i$ and $j$, $f(\cdot)$ is a message function that maps embeddings of the sender and the receiver and the corresponding edge feature to a message vector, $g(\cdot)$ is a node update function that combines the incoming message and the old embedding to be the new node embedding. Both $f(\cdot)$ and $g(\cdot)$ are learnable. Message passing on HMGs is different from that on molecular graphs due to the heterogeneous property of HMGs. Nodes in HMGs are of different orders and they pass messages through edges of different types. A message passing framework needs to learn edge type specific message functions and order specific node update functions to capture this heterogeneous structure. Moreover, the framework should allow inter-order message passing such that the node embeddings can capture information from other orders. For example, by passing messages from $2$-bodies, $1$-bodies can encode edge angle information into their embeddings. Let $i \in \cV_p$ be a node of order $p$ in a HMG and $\mathbf{h}_{p, i}^{(t)}$ be its embedding at the $t$-th layer, we design the message passing framework as
\begin{equation} \label{Eq:HOMP}
\begin{gathered}
    \mathbf{m}_{q, i}^{(t)} = 
    \sum_{j \in \cN_q(i)} f_{qp}^{(t)}\left( \mathbf{h}_{p, i}^{(t)}, \mathbf{h}_{q, j}^{(t)}, \mathbf{e}_{ij} \right) \\
    \mathbf{h}_{p, i}^{(t + 1)} = g_{p}^{(t)} \left(\mathbf{h}_{p, i}^{(t)}, \mathbf{m}_{1, i}^{(t)}, \mathbf{m}_{2,i}^{(t)} , \cdots, \mathbf{m}_{N,i}^{(t)}\right)
\end{gathered}
\end{equation}
where $\cN_q(i)$ the set of nodes of order $q$ that are connected to $i$, $\mathbf{m}_{q, i}^{(t)}$ denotes the aggregated messages from nodes $i$'s neighbor nodes of order $q$, $\mathbf{e}_{ij}$ denotes the edge feature between $i$ and $j$ if they are of the same order, $f_{pq}(\cdot)$ and $g_{p}(\cdot)$ are learnable functions specific to edge type $pq$ and node type (order) $p$, respectively. Compare to the message passing framework on molecular graphs which has two functions to learn, this framework possesses larger model capacity and is able to model many-body interactions explicitly.

\section{Heterogeneous Molecular Graph Neural Networks.}

\begin{figure*}[t]
% \begin{tabular}{cc}
\centering

\includegraphics[width=0.97 \textwidth]{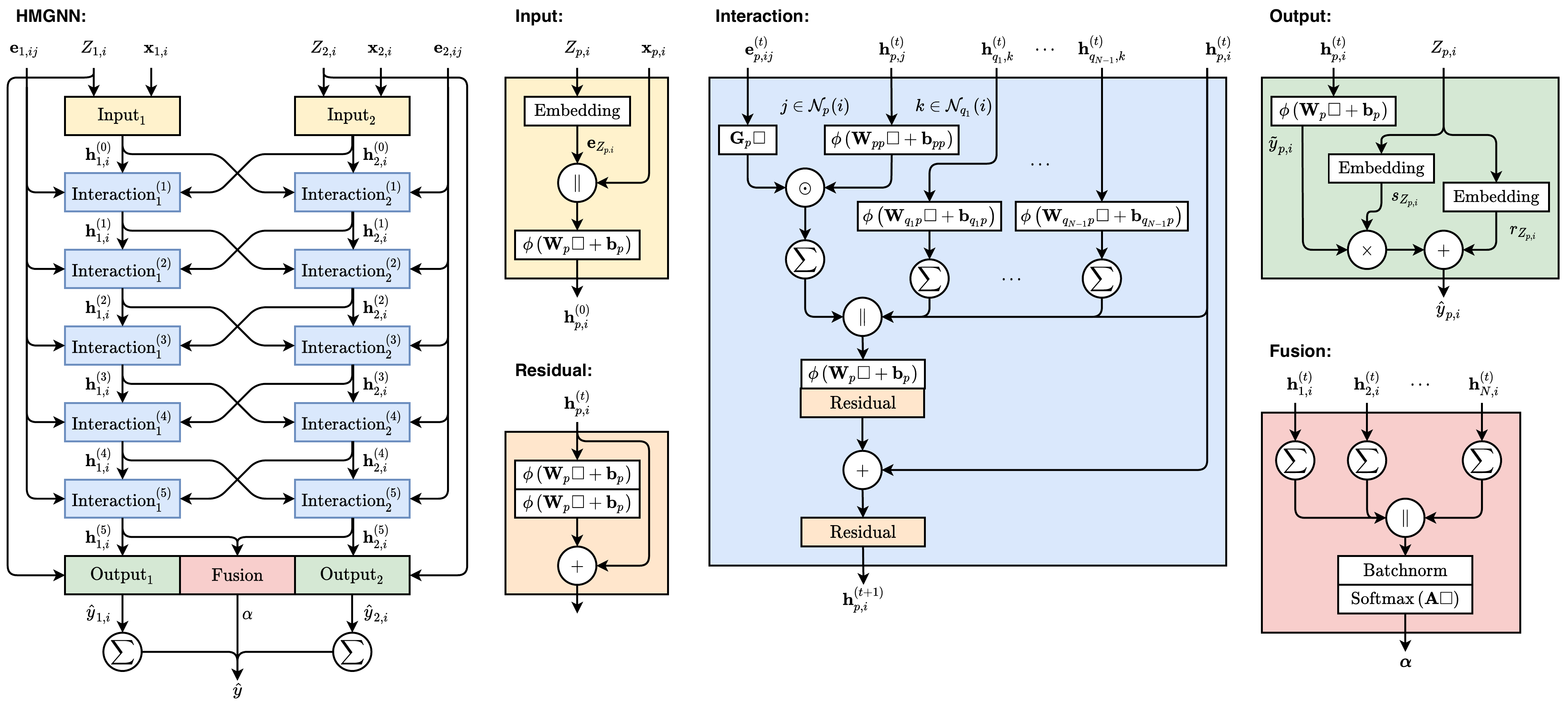}
\caption{Computation flow of heterogeneous molecular graph neural networks (HMGNN) for many-bodies up to order two. We use $\Box$ to represent the input to the function. The activation function is set to be the shifted softplus function, i.e., $\phi(x) = \ln (0.5e^x + 0.5)$. Each many-body order $p$ owns its input module, interaction module, and output module. For each node $i$ of order $p$, an input module converts the discrete and continuous feature of the node to an initial node embedding $\mathbf{h}_{p, i}^{(0)}$. HMGNN passes the initial embeddings through a stack of $T$ interaction modules to encode information from its neighbor nodes of different orders to the node embedding. The outputs of the last interaction module, the final node embedding $\mathbf{h}_{p, i}^{(T + 1)}$, are then fed into a fusion module and an output module to compute a weight vector $\mathbf{\alpha}$ and prediction $\hat{y}_{p, i}$, respectively. HMGNN sums the predictions per many-body order and computes the final prediction as a weighted sum of these summed predictions.}
\label{fig:HMGNN}
\end{figure*}

We present Heterogeneous Molecular Graph Neural Networks (HMGNN) for the purpose of predicting molecule properties. An HMGNN contains four types of modules, input module, interaction module, output module, and fusion module. All the modules except the fusion module are order specific. HMGNNs learn functions for message passing on heterogeneous molecular graphs to compute local node representations, and uses a readout function to combine the representations to form a global molecule representation. HMGNNs compute node-wise contributions to the target property and aggregates them based on their orders. The final prediction is a weighted combination of the predictions of all orders where the weights are computed by an attention mechanism from the global molecule representation. An HMGNN is learned by optimizing a loss function which forces predictions of each order and the fused prediction to be close to the true target. Since the construction of heterogeneous molecular graphs and associated features rely on atom pairwise distances and atomic numbers but not atom coordinates, HMGNNs are invariant under both translations and rotations. HMGNNs are also permutation invariant to atom indices as the message aggregation function in Eq-\ref{Eq:HOMP} and the readout function are permutation invariant \cite{xu2018how}. Figure \ref{fig:HMGNN} shows an overview of the architecture of HMGNN.

% \begin{figure*}[t]
% \centering
% \includegraphics[width=0.9 \textwidth]{BodyNet.pdf}
% \caption{The HomeNet architecture. $\mathbf{x}$ and $\mathbf{e}$ are input node and edge features, respectively. }
% \label{fig:HomeNet}
% \end{figure*}

\subsection{Input Module}

The input module of HMGNN converts raw features of nodes to latent embeddings. As we described in Section \ref{Sec:HMG}, each node $i \in \cV_p$ in a HMG is associated with a discrete feature $Z_{p, i}$ and a continuous feature $\mathbf{x}_{p, i}$. We use an embedding lookup table to map the discrete feature $Z_{p, i}$ to a real value vector $\mathbf{e}_{Z_{p, i}}$ and apply a fully connected layer to the concatenation of the latent vector $\mathbf{e}_{Z_{p, i}}$ and the continuous feature $\mathbf{x}_{p, i}$ to get the initial node embedding
\begin{equation}
    \mathbf{h}_{p, i}^{(1)} = \phi \left( \mathbf{W}^{\text{in}}_p \left(\mathbf{e}_{Z_{p, i}} \parallel \mathbf{x}_{p, i} \right) + \mathbf{b}^{\text{in}}_p \right)
\end{equation}
where $\mathbf{W}^{\text{in}}_p$ and $\mathbf{b}^{\text{in}}_p$ are learnable parameters for nodes of order $p$ ($p$-bodies), $\phi(\cdot)$ is an element-wise activation function, $\parallel$ denotes concatenation of vectors.

\subsection{Interaction Module}

HMGNN stacks $T$ interaction modules to encode information across far reaches of the heterogeneous molecular graph into node embeddings. Each interaction module takes the output embeddings of the previous module and update the embeddings. Note that, edges between nodes of the same orders have features while other edges do not. As a result, we paramatrize the message functions between nodes of the same order as
\begin{equation} \label{eq:mp1}
    \mathbf{m}_{p, i}^{(t)} = \sum_{j \in \cN_p(i)} \mathbf{G}^{(t)}_{p} \mathbf{e}_{p, ij} \odot \phi\left(\mathbf{W}_{pp}^{(t)}\mathbf{h}_{p, j}^{(t)}+\mathbf{b}_{pp}^{(t)}\right)
\end{equation}
and the message functions along edges without features as
\begin{equation} \label{eq:mp2}
    \mathbf{m}_{q, i}^{(t)} = \sum_{j \in \cN_q(i)}  \phi\left(\mathbf{W}_{qp}^{(t)}\mathbf{h}_{q, j}^{(t)}+\mathbf{b}_{qp}^{(t)}\right).
\end{equation}
In Eq-\ref{eq:mp1} and Eq-\ref{eq:mp2}, $\cN_p(i)$ and $\cN_q(i)$ denotes the set of neighbor nodes of order $p$ and order $q$ of node $i$, respectively, $\odot$ denotes the Hadamard product, $\mathbf{G}$, $\mathbf{W}$, and $\mathbf{b}$ are learnable parameters. A node embedding $\mathbf{h}_{p, i}^{(t)}$ is then updated as a function of its old embedding and the incoming messages,
\begin{equation}
    \mathbf{h}_{p, i}^{(t + 1)} = \mathbf{h}_{p, i}^{(t)} +  \phi\left(\mathbf{W}_{p}^{(t)}\left(\mathbf{h}_{p, i}^{(t)} \parallel \mathbf{m}_{1, i}^{(t)}  \parallel \cdots \parallel \mathbf{m}_{N,i}^{(t)}\right) + \mathbf{b}_{p}^{(t)}\right),
\end{equation}
where $\parallel$ denotes concatenation of vectors. The interaction module then refines the node embeddings with two consecutive fully connected layers with residual connections \cite{ResNet}. 

\subsection{Output Module}

Each many-body order $p$ possesses a specific output module that passes the output of its interaction module, final node embeddings $\mathbf{h}_{p, i}^{(T + 1)}$, through a sequence of linear mappings and a aggregation process to compute the estimated value of the target property. First, we use a fully connected layer to convert the node embeddings to node predictions
\begin{equation}
    \Tilde{y}_{p, i} = \mathbf{w}^{\text{out}}_p \mathbf{h}_{p, i}^{(T + 1)} + b^{\text{out}}_p.
\end{equation}
where $\mathbf{w}^{\text{out}}_p$ and $b^{\text{out}}_p$ are learnable parameters for nodes of order $p$. Then we follow \cite{unke2019physnet} and scale the predictions with scaling parameters that are specific to the discrete feature $Z_{p, i}$ of the nodes
\begin{equation}
    \hat{y}_{p, i} =  s_{Z_{p, i}} \Tilde{y}_{p, i} + r_{Z_{p, i}}.
\end{equation}
where $\mathbf{s}$ and $\mathbf{r}$ are learnable embedding lookup tables that map $Z_{p, i}$ to the corresponding scaling factors and shifts. The goal of the scaling layer is to adapt the magnitude of the predictions to different unit systems of the target property.

\subsection{Fusion Module} \label{Sec:fuse}

The fusion module computes a global molecule representation out of the final node embeddings and uses the global representation to weigh the prediction of different orders. We sum the final node embeddings $\mathbf{h}_{p, i}^{(T + 1)}$ of each $p$-body to form an order specific representation and concatenate them to be an intermediate representation
\begin{equation}
    \Tilde{\mathbf{v}} = \sum_{i \in \cV_1} \mathbf{h}_{1, i}^{(T + 1)} \parallel \sum_{i \in \cV_2} \mathbf{h}_{2, i}^{(T + 1)} \parallel \cdots \parallel \sum_{i \in \cV_N} \mathbf{h}_{N, i}^{(T + 1)}.
\end{equation}
Since node embeddings of different orders are computed by different parameters and the number of nodes of the orders also varies, the distributions of the order specific representations could be dramatically different from each other. In order to unify the distributions of the representations and to accelerate training, we apply batch normalization \cite{batchnorm} followed by a fully connected layer on the intermediate representation to obtain the global representation
\begin{equation}
\begin{gathered}
    \mathbf{v} =  \text{BatchNorm}\left(\Tilde{\mathbf{v}}\right) \\
    \mathbf{z} = \phi \left( \mathbf{W} \mathbf{v} + \mathbf{b} \right).
\end{gathered}
\end{equation}
Then we pass the global representation through an attention layer to compute the weight $\alpha_p$ that measures the importance of the predictions of order $p$
\begin{equation}
\alpha_p = \frac{\exp \left( \text{LeakyReLU}\left(\mathbf{z}^T\mathbf{a}_p\right)\right)}{\sum_{q=1}^{N} \exp \left( \text{LeakyReLU}\left(\mathbf{z}^T\mathbf{a}_q\right)\right)}
\end{equation}
where $\mathbf{a}$ are learnable vectors, and $\sum_p \alpha_p = 1$. We can understand the global representation as a query to the knowledge-base distilled in $\mathbf{a}$ for assigning contributions to predictions of different orders. This gives the model better flexibility and explainability in dealing with different molecules.

\subsection{Final prediction}

Inspired by the many-body expansion, we decompose the final prediction as a weighted sum of the prediction of different orders
\begin{equation}
    \hat{y} = \alpha_1 \sum_{i \in \cV_1} \hat{y}_{1, i} + \alpha_2 \sum_{i \in \cV_2} \hat{y}_{2, i} + \alpha_3 \sum_{i \in \cV_3} \hat{y}_{3, i} + \cdots
\end{equation}
where the weights $\alpha_p$ are computed by the fusion module.

\subsection{Model Training}

Since all the modules in HMGNNs except for the fusion module are order specific, and the final prediction is a weighted average of the predictions per order, training HMGNNs by optimizing objective functions that only depend on the final prediction (the fused prediction) may cause gradient vanishing issues for parameters of some orders so that these parameters do not learn enough and lose their prediction utilities. To avoid this issue, we treat the computation of each order as a separate prediction task and propose a multi-task objective function that forces the prediction of all orders together with the final prediction to be close to the true target
\begin{equation} \label{eq:mtlloss}
    \cL = \frac{1}{N + 1} \left(\left\lvert \hat{y} - y \right\rvert + \sum_{p=1}^{N} \left\lvert \hat{y}_p - y \right\rvert \right) + \lambda \left \lVert \Theta \right\rVert^2_2
\end{equation}
where $\hat{y}_p = \sum_{i \in \cV_p} \hat{y}_{p, i}$ is the node order specific prediction, $\Theta$ denotes all trainable parameters of the model, $\lambda \geq 0$ is a hyper-parameter that controls the strength of $L_2$ normalization to prevent the model overfits. This objective function preserves gradient flow for parameters of each order and gives higher training importance to orders that the fussing module assigning larger weights to.

\subsection{Complexity Analysis}

The time and space complexity of HMGNN depends linearly on the number of nodes and edges in a HMG. The number of nodes determines the complexity of the input module and the output module while the number of edges determines the complexity of message passing. 

Let $\cG$ be a molecular graph with $N$ atoms and $H_P(\cG)$ be its HMG that explicitly models up to $P$-bodies. We assume $\cG$ is a complete graph for the worst case scenario. The number of nodes of order $p$ in $H_P(\cG)$ is $\binom{N}{p}$. Let $i \in \cV_p$ be a node of order $p$ (i.e., a $q$-body), $i$ is connected to nodes that are of order $q$ where $q \in \{1, \cdots, P\}$. When $q < p$, the number of $q$ order neighbors of node $i$ is $\binom{p}{q}$ as $i$ is connected to all $q$-bodies who are sub-graphs of $i$; when $q = p$, the number of order $p$ neighbors of $i$ is $p(N - p)$ since $i$ is connected to $p$-bodies who share $p-1$ atoms with $i$; When $q > p$, the number of $q$-body neighbors of $i$ is $\binom{N-p}{q-p}$. As a result, the complexity of message passing is 
\begin{equation}
    \sum_{p = 1}^{P} \binom{N}{p} \left(\sum_{q = 1}^{p - 1} \binom{p}{q} + \sum_{q = p+1}^{P} \binom{N - p}{q - p} + p(N - p) \right)
\end{equation}
and the complexity of the input/output module of HMGNN is $\sum_{p=1}^{P} \binom{N}{p}$.

In this paper, we experiment with HMGs and HMGNNs for up to $2$-bodies, consequently, the time complexity and space complexity of our model are both $\cO(N^3)$. Modern computing architectures such as graphics processing unit (GPU) and tensor processing unit (TPU) are optimized to accelerate this computation. Empirically, HMGNNs can generate property predictions for 10000 randomly drawn molecules from the QM9 dataset in 4 seconds.

\section{Experiments}

We conduct experiments to investigate three research problems in regards of many-body modeling and the HMGNN model
\begin{itemize}
    \item How does HMGNN perform in the molecule property prediction tasks compared against the current state-of-the-art methods?
    \item How does many-body representation, interaction, and prediction contribute to the prediction?
    \item What is the utility of the components of HMGNN?
\end{itemize}

\subsection{Implementation Details} \label{Sec:implement}

We experiment with HMGs and HMGNNs for many-bodies up to order two. There are two types of nodes ($1$-bodies and $2$-bodies), two types of edges with edge features ($1$-$1$ and $2$-$2$ edges), and one type of edge without edge features ($1$-$2$ edges). Since $1$-bodies are atoms, they only have discrete features. Each $2$-body $i$ is determined by its two end atoms and the distance between them $d_{2, i}$. 

There are three types of geometries that we need to model, distance $d_{ij} \in (0, c)$ between $1$-bodies $i$ and $j$, length $l_{2, i} \in (0, c)$ of $2$-bodies, and angle $\theta_{ij} \in [0, \pi]$ between $2$-bodies $i$ and $j$. We use a set of $K$ radial basis functions (RBF) to convert the scalar geometries to real valued vector features. Let $x \in [a, b]$ be a scalar input and $\mathbf{x} \in \cR^K$ be the real valued output of the RBFs, the $k$-th entry of $\mathbf{x}$ is computed as
\begin{equation} \label{eq:RBF}
    x_{k} = \exp{\left( -\beta_k \left( \exp{(-x)} - \mu_k \right)^2 \right)}
\end{equation}
where $\mu_k$ and $\beta_k$ specify the center and width of $x_{k}$. For distance $d_{ij}$ between $1$-bodies, we multiply its feature vector by a continuous monotonic decreasing function $\psi (d_{ij})$ that has $\psi(0) = 1$ and $\psi(c) = 0$. With this formulation, an $1$-body node will have less influence to/from its distant order $1$ neighbors. We follow \cite{unke2019physnet} and set the value of $\mu_k$ to be equally spaced between $\exp{(-a)}$ and $\exp{(-b)}$ while $\beta_k = (2K^{-1}(\exp{(-a)} - \exp{(-b)}))^{-2}$. The goal of using RBFs is to decorrelate the scalar features to accelerate training \cite{SchNet-1}. We apply three different sets of RBFs to convert the distance $d_{ij}$, the length $l_{2, i}$, and the angle $\theta_{ij}$ to the corresponding features $\mathbf{e}_{1, ij}$, $\mathbf{x}_{2, i}$, and $\mathbf{e}_{2, ij}$, respectively. 

We set the latent dimension to be $128$ and use $5$ interaction modules for our experiments. We use the shifted softplus function as the activation function. For ZPVE, $U$, $U_0$, $H$, $G$ and $C_v$, the cutoff distances $c = 3$ while for other targets $c=5$. We initialize the weights of fully connected layers with random orthogonal matrices scaled by the glorot initialization scheme \cite{glorot} and the bias to zero. For learning the parameters of HMGNN, we run the AMSGrad algorithm \cite{amsgrad} with a batch size of $32$ for up to $3000000$ steps and set the $L_2$ regularizer $\lambda$ to be $1 \times 10^{-6}$. We initialize the learning rate to be $1 \times 10^{-3}$ and multiply it with $0.1$ every $2000000$ gradient steps. The training algorithm stops if the MAE on the validation set does not decrease for $1000000$ steps. We implement HMGNN using the Deep Graph Library (DGL) \cite{dgl-1, dgl-2}. 

\subsection{Experimental Setting}

\begin{table}[t]
\captionof{table}{Target properties in the QM9 dataset.}
\begin{center}
\begin{tabular}{l|l}
\toprule
Target & Description \\
\midrule
$\mu$ & Dipole moment \\
$\alpha$ & Isotropic polarizability \\
$\epsilon_{\text{HOMO}}$ & Energy of Highest occupied molecular orbital (HOMO) \\
$\epsilon_{\text{LUMO}}$ & Energy of Lowest occupied molecular orbital (LUMO) \\
$\Delta \epsilon$ & Gap, difference between LUMO and HOMO \\
$ \langle R^2\rangle$ & Electronic spatial extent \\
ZPVE & Zero point vibrational energy \\
$U_0$ & Internal energy at $0$ K \\
$U$ & Internal energy at $298.15$ K \\
$H$ & Enthalpy at $298.15$ K \\
$G$ & Free energy at $298.15$ K \\
$C_v$ & Heat capacity at $298.15$ K \\
\bottomrule
\end{tabular}
\label{Table:Property}
\end{center}
\end{table}

We evaluate the performance of the proposed model on the QM9 dataset \cite{QM9-1, QM9-2}. QM9 is a widely used benchmark for evaluating models that predict molecule properties. It consists of around $130$K equilibrium molecules associated with $12$ geometric, energetic, electronic, and thermodynamic properties. The properties are described in Table \ref{Table:Property}. These molecules contain up to nine heavy atoms (C, O, N, and F). We randomly select $110000$ molecules for training, $10000$ molecules for validation, and $10831$ molecules as the test set. We conduct model selection for different targets on the validation set and report the mean absolute error (MAE) of the best performing models. For properties with atomic reference values ($U_0$, $U$, $H$, $G$, $C_v$), we subtract the original value by the per-atom-type reference values to be the target. Since $\Delta \epsilon$ is defined as the gap between $\epsilon_{\text{LUMO}}$ and $\epsilon_{\text{HOMO}}$, we predict it as $\Delta \epsilon = \epsilon_{\text{LUMO}} - \epsilon_{\text{HOMO}}$. In our experiments, we convert the units of $\epsilon_{\text{HOMO}}$, $\epsilon_{\text{LUMO}}$, $\Delta \epsilon$, ZPVE, $U_0$, $U$, $H$, $G$ to eV.

We compare the performance of HMGNN with six state-of-the-art methods, enn-s2s \cite{MPNN}, SchNet \cite{SchNet-1}, neural message passing with edge updates (NMP-edge) \cite{NMP-edge}, Cormorant \cite{anderson2019cormorant}, PhysNet \cite{unke2019physnet}, and directional message passing neural network (DimeNet) \cite{DimeNet}. Results of enn-s2s, SchNet, NMP-edge, Cormorant, and DimeNet are from the corresponding papers. We take the results of PhysNet from \cite{DimeNet}.

\subsection{Prediction Performance}

\begin{table*}[t]
\captionof{table}{Mean absolute error on QM9 with 110K training molecules. In each row, we use boldface for the best performance method. Column HMGNN-$1$ and HMGNN-$2$ correspond to the performance of summing over predictions of $1$-bodies and $2$-bodies, respectively.}
\begin{center}
\begin{tabular}{l|l|rrrrrr|rrr}
\toprule
Target & Unit & enn-s2s & SchNet & NMP-edge & Cormorant & PhysNet & DimeNet & HMGNN-$1$ & HMGNN-$2$ & HMGNN \\
\midrule
$\mu$ & D & 0.030 & 0.033 & 0.029 & 0.038 & 0.0529 & 0.0286 & 0.0276	& 0.0283 & \textbf{0.0272} \\
$\alpha$ & $a_0^3$ & 0.092 & 0.235 & 0.077 & 0.085 & 0.0615 & \textbf{0.0469} & 0.0571 & 0.0647 & 0.0561 \\
$\epsilon_{\text{HOMO}}$ &  meV & 43 & 41 & 36.7 & 34 & 32.9 & 27.8 & 24.94 & 26.31 & \textbf{24.78}  \\
$\epsilon_{\text{LUMO}}$ & meV & 37 & 34 & 30.8 & 38 & 27.4 & \textbf{19.7} & 20.72 & 21.42 & 20.61 \\
$\Delta \epsilon$ &  meV & 69 & 63 & 58.0 & 61 & 42.5 & 34.8 & 33.44  & 35.02 & \textbf{33.31} \\
$ \langle R^2\rangle$ &  $a_0^2$ & 0.180 & 0.073 & \textbf{0.072} & 0.961 & 0.765 & 0.331 & 0.43 & 0.6 & 0.416 \\
ZPVE & meV & 1.5 & 1.7 & 1.49 & 2.03 & 1.39 & 1.29 & 1.24 & 1.34 &	\textbf{1.18} \\
$U_0$ & meV & 19 & 14 & 10.5 & 22 & 8.15 & 8.02 & 6.19 & 9.06 & \textbf{5.92} \\
$U$ & meV & 19 & 19 & 10.6 & 21 & 8.34 & 7.89 & 7.22 & 11 & \textbf{6.85} \\
$H$ & meV & 17 & 14 & 11.3 & 21 & 8.42 & 8.11 & 6.35 & 8.37 & \textbf{6.08} \\
$G$ & meV & 19 & 14 & 12.2 & 20 & 9.40 & 8.98 & 7.95 & 11.06 & \textbf{7.61} \\
$c_v$ & $\frac{\text{cal}}{\text{mol K}}$ & 0.040 & 0.033 & 0.032 & 0.026 & 0.0280 & 0.0249 & 0.0241 & 0.025 & \textbf{0.0233} \\
\bottomrule
\end{tabular}
\label{Table:Result}
\end{center}
\end{table*}

\begin{figure*}[t]
% \begin{tabular}{cc}
\centering
\subfigure[$U_0$.]{
\includegraphics[width=0.23 \textwidth]{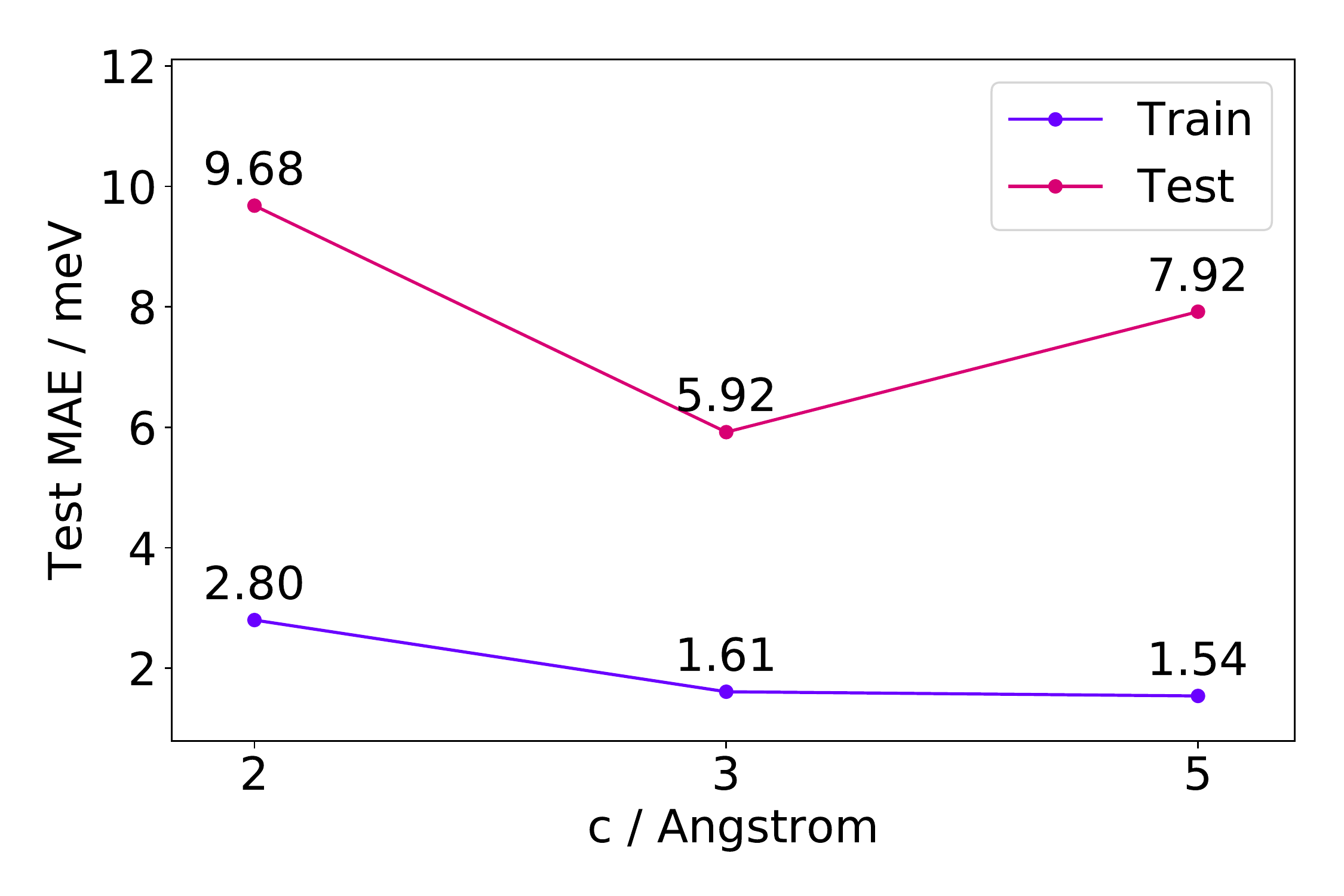}
\label{subfig:U0}}
\subfigure[$C_v$.]{
\includegraphics[width=0.23 \textwidth]{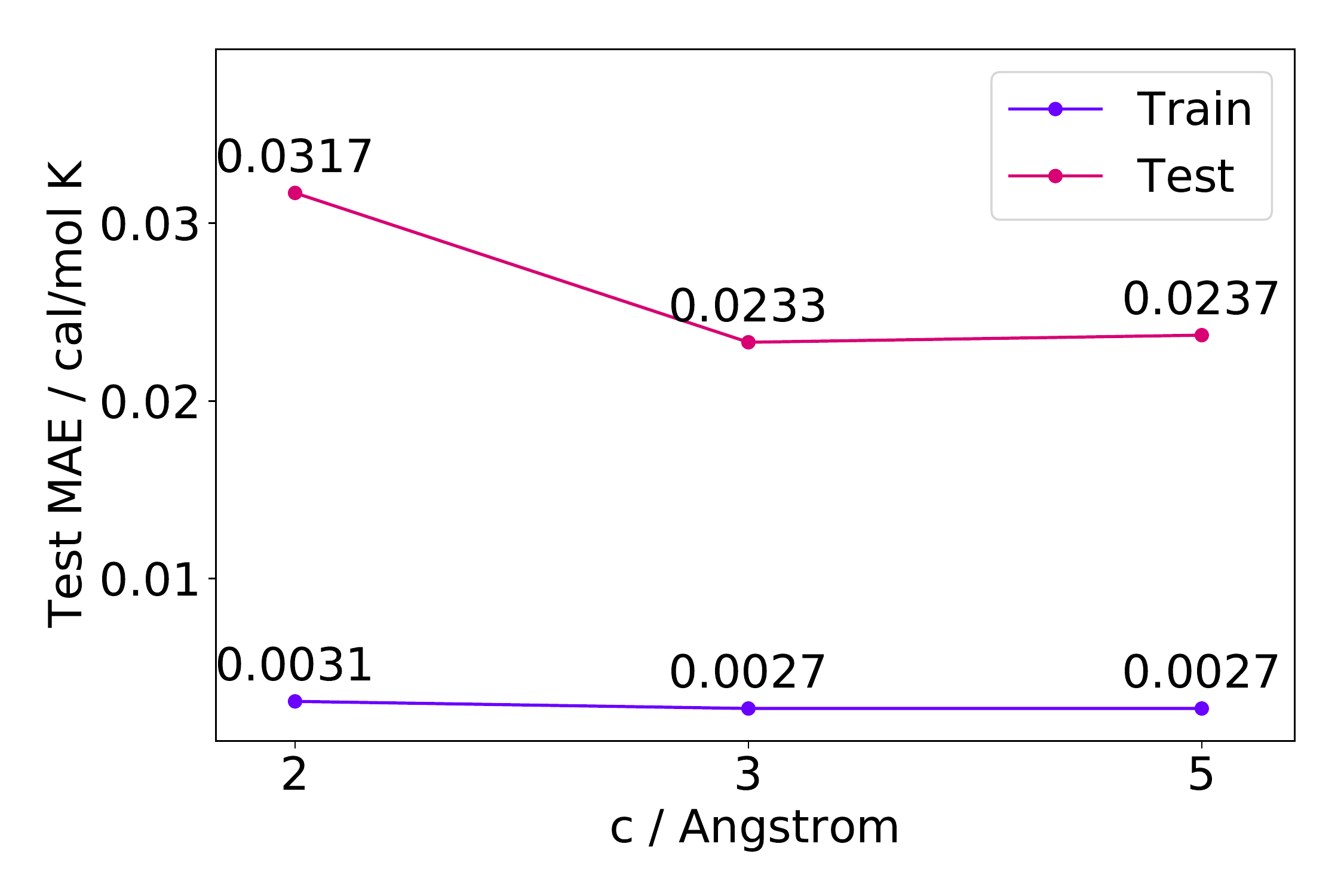}
\label{subfig:Cv}}
\subfigure[ZPVE.]{
\includegraphics[width=0.23 \textwidth]{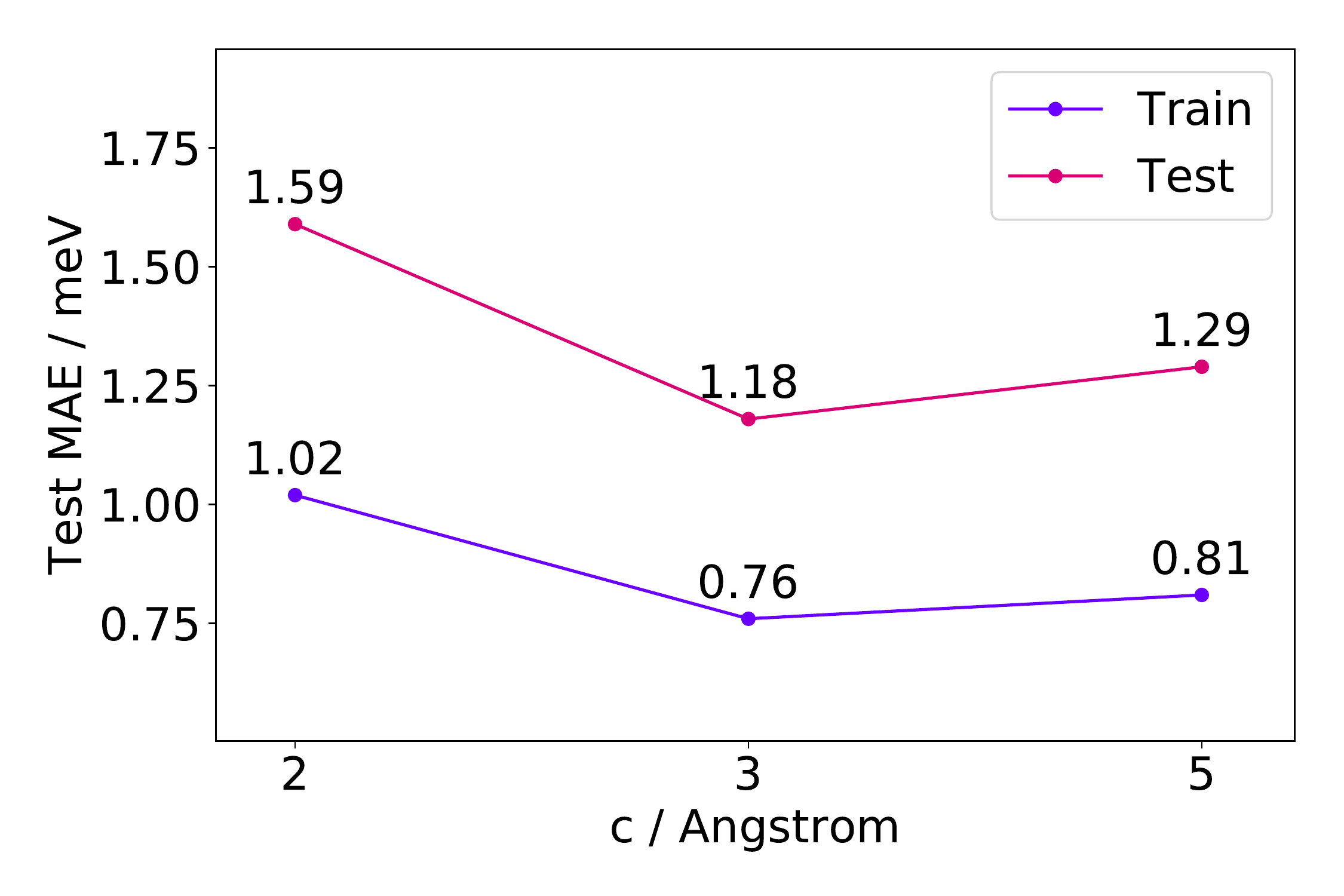}
\label{subfig:zpve}}
\subfigure[$\mu$.]{
\includegraphics[width=0.23 \textwidth]{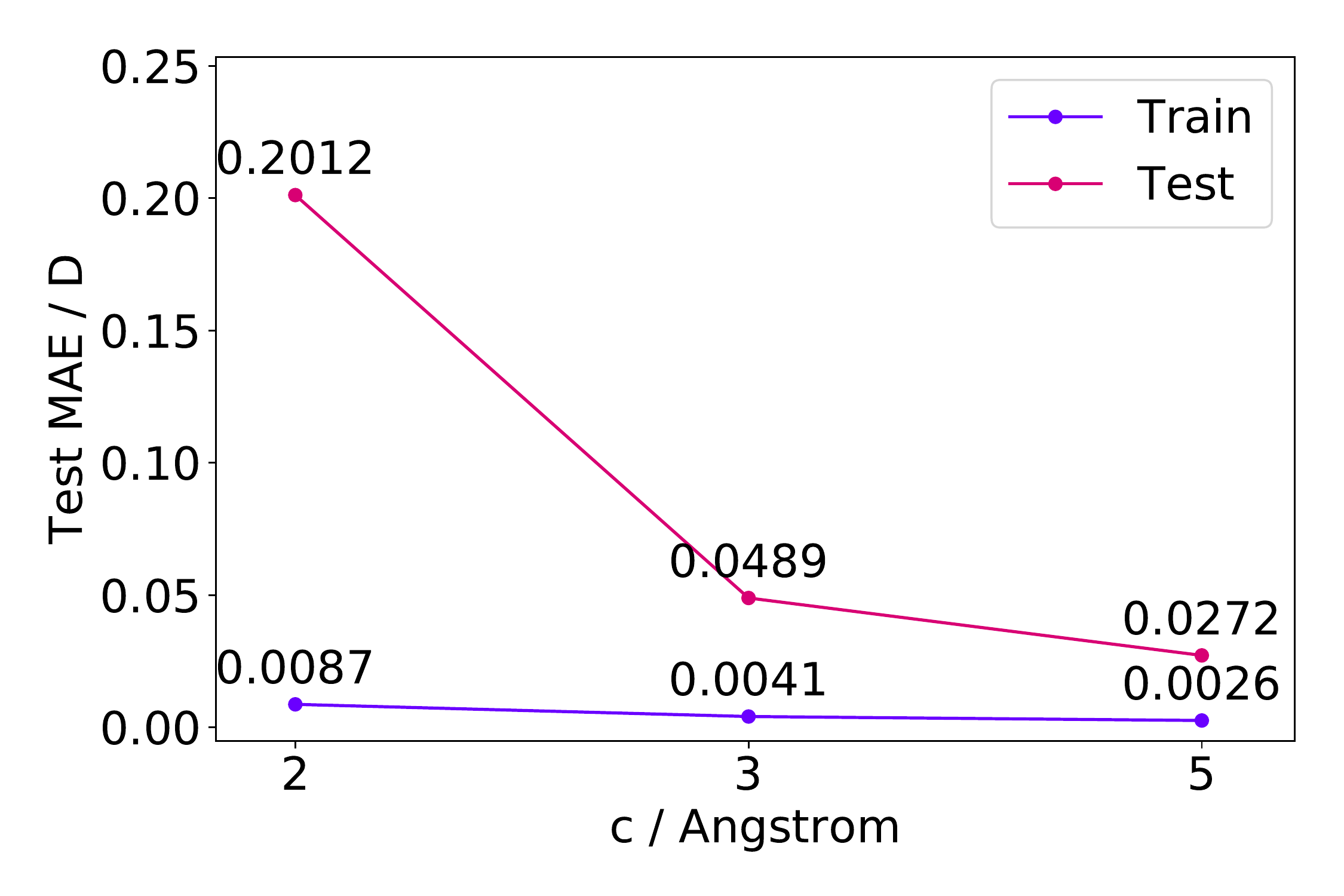}
\label{subfig:mu}}

\caption{Effect of the cutoff distance $c$ on prediction performance on four target properties. }
\label{fig:cut_r}
\end{figure*}

We show the prediction performance of HMGNN and the competing methods on the 12 properties of QM9 in Table \ref{Table:Result}. Our proposed method sets the new state-of-the-art on 9 out of the 12 target properties. HMGNN's performance aligns with the best results on the remaining targets with an exception of $\langle R^2\rangle$. We also present the performance of summing over predictions over $1$-bodies (HMGNN-$1$) and $2$-bodies (HMGNN-$2$), respectively. Although the performance of HMGNN-$2$ is consistently worse than HMGNN-$1$, their weighted combination outperforms any of the standalone prediction. This demonstrates the effectiveness of the fusion module driven by the global molecule representations and the attention mechanism, and that explicitly modeling and computing predictions of many-bodies can be beneficial for chemical prediction tasks.

We analyze the effect of a critical hyper-parameter, the cutoff distance $c$, on prediction performances of four types of properties. We choose $U_0$ to represent properties related to atomization energies ($U_0$, $U$, $H$, $G$), $C_v$ to represent thermodynamic properties ($C_v$), ZPVE to represent properties related to fundamental vibrations of the molecule (ZPVE), and $\mu$ to represent electronic properties ($\mu$, $\alpha$, $\epsilon_{\text{HOMO}}$, $\epsilon_{\text{LUMO}}$, $\Delta \epsilon$, $ \langle R^2\rangle$) \cite{MPNN}. We present the training and test mean absolute error (MAE) of HMGNNs on HMGs constructed with $c \in \left\{2, 3, 5\right\}$ in Figure \ref{fig:cut_r}. 

When constructing molecular graphs as well as HMGs, the larger the cutoff distance we choose, the less geometric information about the molecules that we lose. However, a large cutoff value does not always lead to better performance. In Figure \ref{fig:cut_r}, despite the training error decreases across all the four targets as the cutoff value increases, the test error shows an increasing trend for three properties. This is a signal that the model over-fits the training set on the three properties. This is because of the large model capacity of HMGNNs as they have one set of parameters for each many-body order. An HMGNN of order $N$ possesses $N$ times the number of parameters of a normal GNN-based model. 

% The one property $\mu$ that benefits from a large cut off value is the norm of the dipole moment which relates to the electric field far from a molecule. As a result, predicting $\mu$ requires modeling long-range interactions. In contrast, the other three targets are relatively local as they achieves the best performance with interactions of shorter range. As HMGNN has a large capacity, it is important to pick the right bias for it to perform well.

\subsection{Ablation Study}

\begin{table}[t]
\captionof{table}{Ablation study on $U_0$ and $C_v$.}
\begin{center}
\begin{tabular}{llccc}
\toprule
Target & Architecture & HMGNN-1 & HMGNN-2 & HMGNN \\
\midrule
\multirow{4}{*}{$U_0$} & Default & 6.19 & 9.06 & 5.92 \\
& Remove MTL & 8.22 & 9716.95 & 8.22 \\
& Remove IOMP & 10.26 & 8.18 & 7.88 \\
& Remove HO & 10.08 & - &  - \\
\midrule
\multirow{4}{*}{$C_v$} & Default & 0.0241 & 0.0250 & 0.0233 \\
& Remove MTL & 0.0247 & 1.4022 & 0.0247\\
& Remove IOMP & 0.0297 & 0.0275 & 0.0244 \\
& Remove HO & 0.0289 & - &  - \\
\bottomrule
\end{tabular}
\label{Table:mtl}
\end{center}
\end{table}

In this section, we conduct ablation study on two targets (i.e., $U_0$, $C_v$) to demonstrate the importance of the multi-task learning loss, inter-order message passing, and explicit modeling of high-order bodies in improving the performance of molecular property prediction. We propose three variants of the HMGNN model and show their results in Table \ref{Table:mtl}.

\subsubsection{Remove MTL (Multi-Task Learning loss)} This variant has the same specification with the default model. It differs with the default model in that it is trained by minimizing the naive loss $\lvert \hat{y} - y \rvert$ instead of the multi-task learning loss that we proposed in Eq-\ref{eq:mtlloss}. As shown in Table \ref{Table:mtl}, the $2$-bodies of this variant lose their prediction power while the fusion module gives all attention weights to the $1$-bodies, and as a result, the performance of this variant is worse than the default HMGNN. Furthermore, the prediction of $1$-bodies (i.e., HMGNN-1) is also less accurate than the default model. 

\subsubsection{Remove IOMP (Inter-Order Message Passing)} This variant removes edges/messages between $1$-bodies and $2$-bodies, as a result, information of the two orders are not shared. We can see that the performance of HMGNN-$1$ and HMGNN drops in the prediction of both $U_0$ and $C_v$. This demonstrates the importance of inter-order message passing. However, the prediction accuracy of HMGNN-$2$ on $U_0$ is better than models with inter-order message passing. This might because $2$-bodies (both distance and angle) contain more geometric information than $1$-bodies (only distance). 

\subsubsection{Remove HO (High-Order modeling)} This variant removes high-order related modeling ($2$-body interaction, representation, and prediction) and is similar to existing GNN-based prediction methods (i.e., PhysNet). As shown in Table \ref{Table:mtl}, this method performs worse than HMGNN-1 of the variant that removes multi-task learning loss. This shows another evidence of the effectiveness of inter-order message passing.

\subsection{Visualization of Attention weights}

\begin{figure*}[t]
% \begin{tabular}{cc}
\centering
\subfigure[$U_0$.]{
\includegraphics[width=0.33 \textwidth]{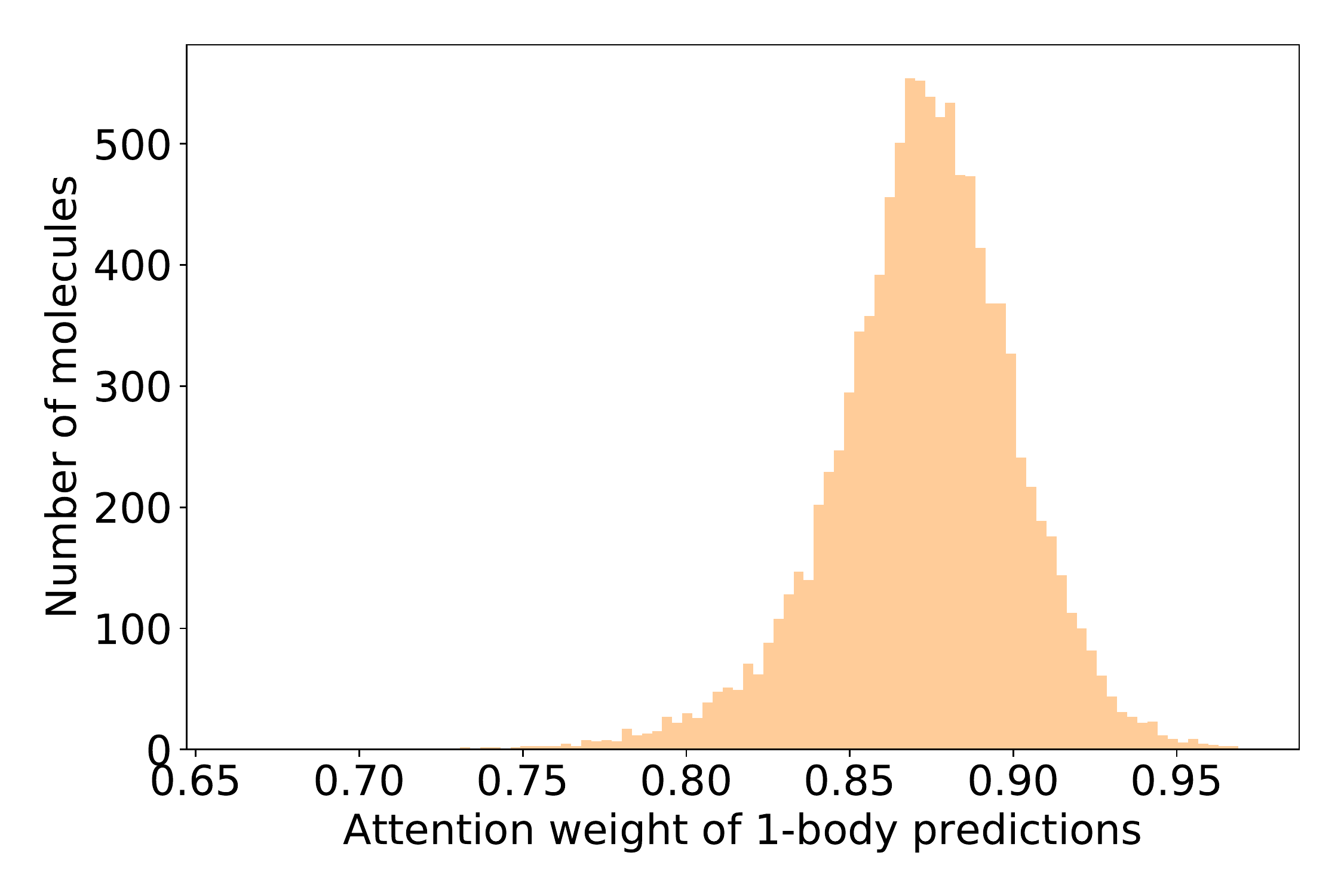}
\label{subfig:U0}}
\subfigure[$C_v$.]{
\includegraphics[width=0.33 \textwidth]{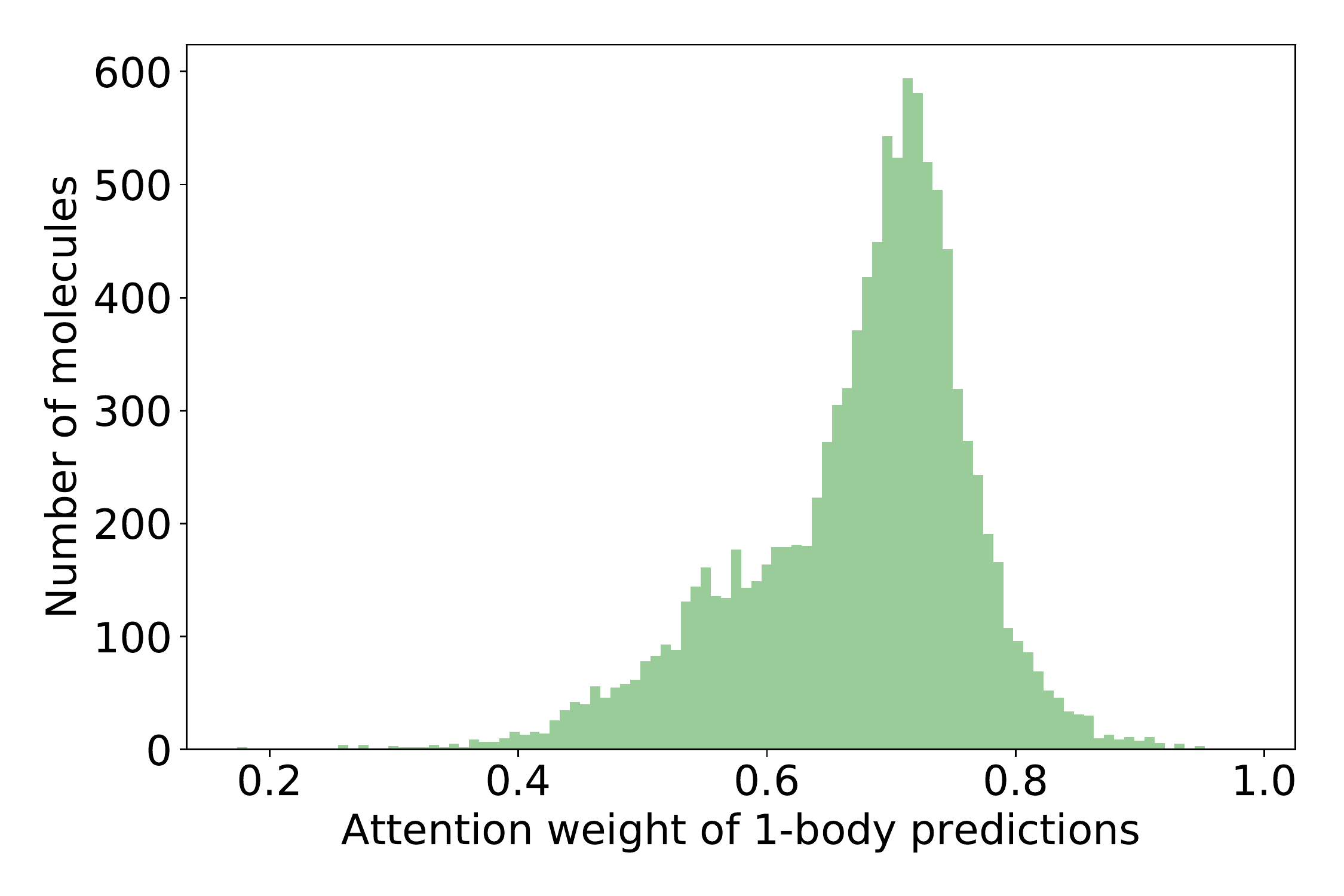}
\label{subfig:Cv}}
\subfigure[ZPVE.]{
\includegraphics[width=0.33 \textwidth]{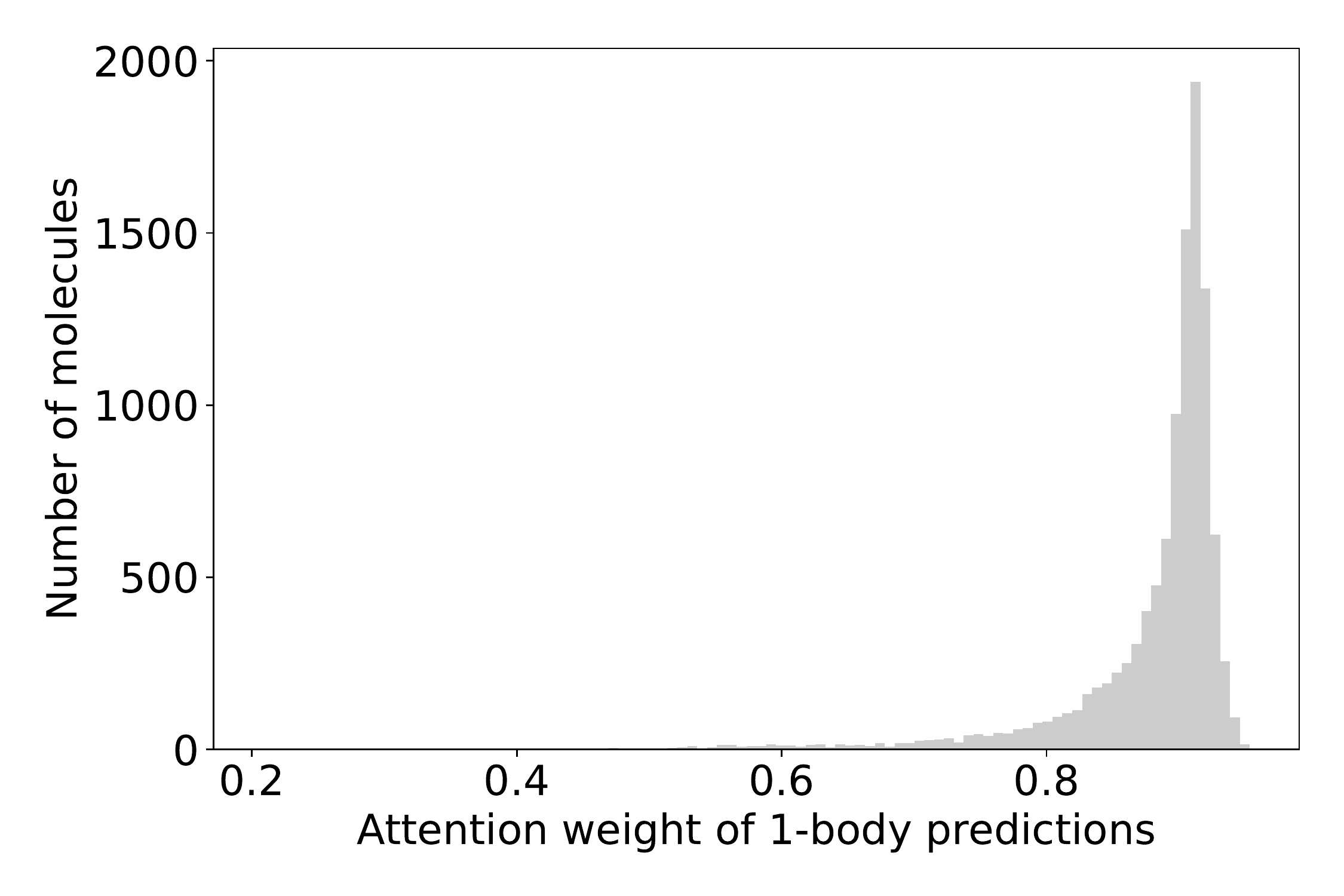}
\label{subfig:zpve}}
\subfigure[$\mu$.]{
\includegraphics[width=0.33 \textwidth]{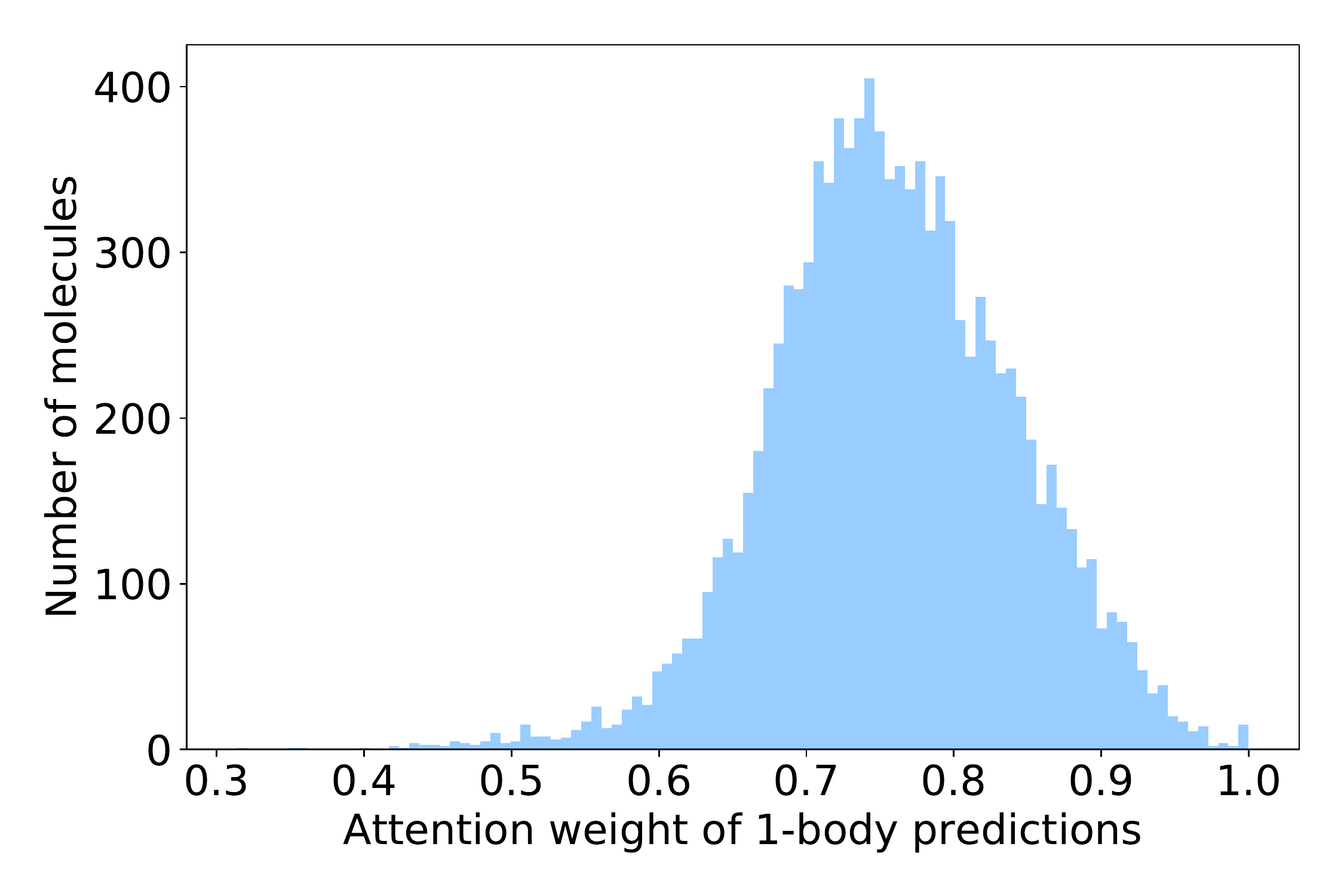}
\label{subfig:mu}}

\caption{Attention weights generate by the fusion module for predicting the four properties. }
\label{fig:att}
\end{figure*}

In Figure \ref{fig:att}, we show the attention scores of the $1$-body predictions generated by the fusion module for predicting $U_0$, $C_v$, $\mu$, and ZPVE on the test set. Since we only experiment with many-bodies up to the second order, the attention weights of the $2$-bodies is one minus that of the $1$-bodies. On the four types of chemical properties, $1$-body contribution dominates the prediction of most of the molecules. However, $2$-body predictions also take a considerable amount of attention. 

\section{Conclusion}

We propose a novel heterogeneous graph based molecule representation, heterogeneous molecular graph (HMG), to model many-body representations and interactions. Inspired by the many-body expansion of energy surfaces, we design a heterogeneous molecular graph neural network (HMGNN) to leverage the rich information stored in HMGs for molecular prediction tasks. HMGNN follows a message passing paradigm and leverages global molecule representations using an attention mechanism. We propose to train HMGNNs by optimizing a multi-task learning loss. HMGNN achieves state-of-the-art performance on 9 out of 12 properties on the QM9 dataset. Experiments also show that the multi-task learning loss improves the generalization of the model. In this paper, we only model many-bodies up to the second order, future works should aim to model many-bodies of higher than third orders and also to enable HMGNNs for another important chemical prediction tasks, molecular dynamics simulations.

\section{Acknowledgement}

This work was supported in part by NSF (1447788, 1704074, 1757916, 1834251), Army
Research Office (W911NF1810344), Intel Corp, and the Digital Technology Center at the
University of Minnesota. Access to research and computing facilities was provided by
the Digital Technology Center and the Minnesota Supercomputing Institute. We are grateful to Mingjian Wen for his fruitful comments, corrections and inspiration.

\bibliographystyle{IEEEtran}
\bibliography{conference_101719}

\begin{thebibliography}{10}
\providecommand{\url}[1]{#1}
\csname url@samestyle\endcsname
\providecommand{\newblock}{\relax}
\providecommand{\bibinfo}[2]{#2}
\providecommand{\BIBentrySTDinterwordspacing}{\spaceskip=0pt\relax}
\providecommand{\BIBentryALTinterwordstretchfactor}{4}
\providecommand{\BIBentryALTinterwordspacing}{\spaceskip=\fontdimen2\font plus
\BIBentryALTinterwordstretchfactor\fontdimen3\font minus
  \fontdimen4\font\relax}
\providecommand{\BIBforeignlanguage}[2]{{%
\expandafter\ifx\csname l@#1\endcsname\relax
\typeout{** WARNING: IEEEtran.bst: No hyphenation pattern has been}%
\typeout{** loaded for the language `#1'. Using the pattern for}%
\typeout{** the default language instead.}%
\else
\language=\csname l@#1\endcsname
\fi
#2}}
\providecommand{\BIBdecl}{\relax}
\BIBdecl

\bibitem{hohenberg1964inhomogeneous}
P.~Hohenberg and W.~Kohn, ``Inhomogeneous electron gas,'' \emph{Physical
  review}, vol. 136, no.~3B, p. B864, 1964.

\bibitem{unke2019physnet}
O.~T. Unke and M.~Meuwly, ``Physnet: a neural network for predicting energies,
  forces, dipole moments, and partial charges,'' \emph{Journal of chemical
  theory and computation}, vol.~15, no.~6, pp. 3678--3693, 2019.

\bibitem{SchNet-1}
K.~Sch{\"u}tt, P.-J. Kindermans, H.~E.~S. Felix, S.~Chmiela, A.~Tkatchenko, and
  K.-R. M{\"u}ller, ``Schnet: A continuous-filter convolutional neural network
  for modeling quantum interactions,'' in \emph{Advances in neural information
  processing systems}, 2017, pp. 991--1001.

\bibitem{SchNet-2}
K.~T. Sch{\"u}tt, H.~E. Sauceda, P.-J. Kindermans, A.~Tkatchenko, and K.-R.
  M{\"u}ller, ``Schnet--a deep learning architecture for molecules and
  materials,'' \emph{The Journal of Chemical Physics}, vol. 148, no.~24, p.
  241722, 2018.

\bibitem{MGCN}
C.~Lu, Q.~Liu, C.~Wang, Z.~Huang, P.~Lin, and L.~He, ``Molecular property
  prediction: A multilevel quantum interactions modeling perspective,'' in
  \emph{Proceedings of the AAAI Conference on Artificial Intelligence},
  vol.~33, 2019, pp. 1052--1060.

\bibitem{MPNN}
J.~Gilmer, S.~S. Schoenholz, P.~F. Riley, O.~Vinyals, and G.~E. Dahl, ``Neural
  message passing for quantum chemistry,'' in \emph{Proceedings of the 34th
  International Conference on Machine Learning-Volume 70}.\hskip 1em plus 0.5em
  minus 0.4em\relax JMLR. org, 2017, pp. 1263--1272.

\bibitem{NMP-edge}
P.~B. J{\o}rgensen, K.~W. Jacobsen, and M.~N. Schmidt, ``Neural message passing
  with edge updates for predicting properties of molecules and materials,''
  \emph{arXiv preprint arXiv:1806.03146}, 2018.

\bibitem{MBE-1}
F.~H. Stillinger and T.~A. Weber, ``Computer simulation of local order in
  condensed phases of silicon,'' \emph{Physical review B}, vol.~31, no.~8, p.
  5262, 1985.

\bibitem{MBE-2}
M.~J. Elrod and R.~J. Saykally, ``Many-body effects in intermolecular forces,''
  \emph{Chemical reviews}, vol.~94, no.~7, pp. 1975--1997, 1994.

\bibitem{MBE-NN}
K.~Yao, J.~E. Herr, and J.~Parkhill, ``The many-body expansion combined with
  neural networks,'' \emph{The Journal of chemical physics}, vol. 146, no.~1,
  p. 014106, 2017.

\bibitem{MTL}
S.~Ruder, ``An overview of multi-task learning in deep neural networks,''
  \emph{arXiv preprint arXiv:1706.05098}, 2017.

\bibitem{QM9-1}
L.~Ruddigkeit, R.~Van~Deursen, L.~C. Blum, and J.-L. Reymond, ``Enumeration of
  166 billion organic small molecules in the chemical universe database
  gdb-17,'' \emph{Journal of chemical information and modeling}, vol.~52,
  no.~11, pp. 2864--2875, 2012.

\bibitem{QM9-2}
R.~Ramakrishnan, P.~O. Dral, M.~Rupp, and O.~A. Von~Lilienfeld, ``Quantum
  chemistry structures and properties of 134 kilo molecules,'' \emph{Scientific
  data}, vol.~1, p. 140022, 2014.

\bibitem{DFT}
R.~G. Parr, ``Density functional theory of atoms and molecules,'' in
  \emph{Horizons of Quantum Chemistry}.\hskip 1em plus 0.5em minus 0.4em\relax
  Springer, 1980, pp. 5--15.

\bibitem{faber2017machine}
F.~A. Faber, L.~Hutchison, B.~Huang, J.~Gilmer, S.~S. Schoenholz, G.~E. Dahl,
  O.~Vinyals, S.~Kearnes, P.~F. Riley, and O.~A. von Lilienfeld, ``Machine
  learning prediction errors better than dft accuracy,'' \emph{arXiv preprint
  arXiv:1702.05532}, 2017.

\bibitem{bartok2017machine}
A.~P. Bart{\'o}k, S.~De, C.~Poelking, N.~Bernstein, J.~R. Kermode,
  G.~Cs{\'a}nyi, and M.~Ceriotti, ``Machine learning unifies the modeling of
  materials and molecules,'' \emph{Science advances}, vol.~3, no.~12, p.
  e1701816, 2017.

\bibitem{GCN}
T.~N. Kipf and M.~Welling, ``Semi-supervised classification with graph
  convolutional networks,'' \emph{arXiv preprint arXiv:1609.02907}, 2016.

\bibitem{GAT}
\BIBentryALTinterwordspacing
P.~Veličković, G.~Cucurull, A.~Casanova, A.~Romero, P.~Liò, and Y.~Bengio,
  ``Graph attention networks,'' in \emph{International Conference on Learning
  Representations}, 2018. [Online]. Available:
  \url{https://openreview.net/forum?id=rJXMpikCZ}
\BIBentrySTDinterwordspacing

\bibitem{SAGE}
W.~Hamilton, Z.~Ying, and J.~Leskovec, ``Inductive representation learning on
  large graphs,'' in \emph{Advances in neural information processing systems},
  2017, pp. 1024--1034.

\bibitem{DiffPool}
Z.~Ying, J.~You, C.~Morris, X.~Ren, W.~Hamilton, and J.~Leskovec,
  ``Hierarchical graph representation learning with differentiable pooling,''
  in \emph{Advances in neural information processing systems}, 2018, pp.
  4800--4810.

\bibitem{DTNN}
K.~T. Sch{\"u}tt, F.~Arbabzadah, S.~Chmiela, K.~R. M{\"u}ller, and
  A.~Tkatchenko, ``Quantum-chemical insights from deep tensor neural
  networks,'' \emph{Nature communications}, vol.~8, no.~1, pp. 1--8, 2017.

\bibitem{HIP-NN}
N.~Lubbers, J.~S. Smith, and K.~Barros, ``Hierarchical modeling of molecular
  energies using a deep neural network,'' \emph{The Journal of chemical
  physics}, vol. 148, no.~24, p. 241715, 2018.

\bibitem{DimeNet}
\BIBentryALTinterwordspacing
J.~Klicpera, J.~Groß, and S.~Günnemann, ``Directional message passing for
  molecular graphs,'' in \emph{International Conference on Learning
  Representations}, 2020. [Online]. Available:
  \url{https://openreview.net/forum?id=B1eWbxStPH}
\BIBentrySTDinterwordspacing

\bibitem{cohen2016group}
T.~Cohen and M.~Welling, ``Group equivariant convolutional networks,'' in
  \emph{International conference on machine learning}, 2016, pp. 2990--2999.

\bibitem{anderson2019cormorant}
B.~Anderson, T.~S. Hy, and R.~Kondor, ``Cormorant: Covariant molecular neural
  networks,'' in \emph{Advances in Neural Information Processing Systems},
  2019, pp. 14\,510--14\,519.

\bibitem{thomas2018tensor}
N.~Thomas, T.~Smidt, S.~Kearnes, L.~Yang, L.~Li, K.~Kohlhoff, and P.~Riley,
  ``Tensor field networks: Rotation-and translation-equivariant neural networks
  for 3d point clouds,'' \emph{arXiv preprint arXiv:1802.08219}, 2018.

\bibitem{kondor2018covariant}
\BIBentryALTinterwordspacing
R.~Kondor, T.~S. Hy, H.~Pan, B.~M. Anderson, and S.~Trivedi, ``Covariant
  compositional networks for learning graphs,'' 2018. [Online]. Available:
  \url{https://openreview.net/forum?id=S1TgE7WR-}
\BIBentrySTDinterwordspacing

\bibitem{xu2018how}
\BIBentryALTinterwordspacing
K.~Xu, W.~Hu, J.~Leskovec, and S.~Jegelka, ``How powerful are graph neural
  networks?'' in \emph{International Conference on Learning Representations},
  2019. [Online]. Available: \url{https://openreview.net/forum?id=ryGs6iA5Km}
\BIBentrySTDinterwordspacing

\bibitem{ResNet}
K.~He, X.~Zhang, S.~Ren, and J.~Sun, ``Deep residual learning for image
  recognition,'' in \emph{Proceedings of the IEEE conference on computer vision
  and pattern recognition}, 2016, pp. 770--778.

\bibitem{batchnorm}
S.~Ioffe and C.~Szegedy, ``Batch normalization: Accelerating deep network
  training by reducing internal covariate shift,'' \emph{arXiv preprint
  arXiv:1502.03167}, 2015.

\bibitem{glorot}
X.~Glorot and Y.~Bengio, ``Understanding the difficulty of training deep
  feedforward neural networks,'' in \emph{Proceedings of the thirteenth
  international conference on artificial intelligence and statistics}, 2010,
  pp. 249--256.

\bibitem{amsgrad}
\BIBentryALTinterwordspacing
S.~J. Reddi, S.~Kale, and S.~Kumar, ``On the convergence of adam and beyond,''
  in \emph{International Conference on Learning Representations}, 2018.
  [Online]. Available: \url{https://openreview.net/forum?id=ryQu7f-RZ}
\BIBentrySTDinterwordspacing

\bibitem{dgl-1}
D.~Zheng, M.~Wang, Q.~Gan, Z.~Zhang, and G.~Karypis, ``Learning graph neural
  networks with deep graph library,'' in \emph{Companion Proceedings of the Web
  Conference 2020}, 2020, pp. 305--306.

\bibitem{dgl-2}
M.~Wang, L.~Yu, D.~Zheng, Q.~Gan, Y.~Gai, Z.~Ye, M.~Li, J.~Zhou, Q.~Huang,
  C.~Ma \emph{et~al.}, ``Deep graph library: Towards efficient and scalable
  deep learning on graphs,'' \emph{arXiv preprint arXiv:1909.01315}, 2019.

\end{thebibliography}

\end{document}